\documentclass[nolineno]{ieeeaccess}
\usepackage[nocompress]{cite}
\usepackage{amsmath,amssymb,amsfonts,amsthm,booktabs,xurl}
\usepackage{graphicx,textcomp,bm,etoolbox,array}
\usepackage{placeins,needspace}
\usepackage[hidelinks,bookmarks=false]{hyperref}
\hypersetup{pdftitle={Exact Fusion of Softmax Attention Heads Can Require Exponentially Many Atoms},pdfauthor={Hiroyasu Usami, Keisuke Hara, Ayato Tsuboi, Naohiko Matsuda},pdfsubject={Research preprint},hidelinks}
\makeatletter
\AtBeginDocument{\DeclareMathVersion{bold}
\SetSymbolFont{operators}{bold}{T1}{times}{b}{n}
\SetSymbolFont{NewLetters}{bold}{T1}{times}{b}{it}
\SetMathAlphabet{\mathrm}{bold}{T1}{times}{b}{n}
\SetMathAlphabet{\mathit}{bold}{T1}{times}{b}{it}
\SetMathAlphabet{\mathbf}{bold}{T1}{times}{b}{n}
\SetMathAlphabet{\mathtt}{bold}{OT1}{pcr}{b}{n}
\SetSymbolFont{symbols}{bold}{OMS}{cmsy}{b}{n}
\renewcommand\boldmath{\@nomath\boldmath\mathversion{bold}}}
\patchcmd{\@maketitle}{\vss\doifont Digital Object Identifier\space\@doi\vss}{}{}{\PackageError{access-migration}{DOI patch failed}{}}
\let\access@titlepage\ps@titlepage
\def\ps@titlepage{\access@titlepage
  \def\@oddfoot{\hfil{\footerpagefont\thepage}}
  \def\@evenfoot{{\footerpagefont\thepage}\hfil}}
\let\access@headings\ps@headings
\def\ps@headings{\access@headings
  \def\@oddfoot{\hfil{\footerpagefont\thepage}}
  \def\@evenfoot{{\footerpagefont\thepage}\hfil}}
\ps@headings
\patchcmd{\IEEEbiographynophoto}{\vskip 4\baselineskip plus 1fil minus 0\baselineskip}{\vskip 2\baselineskip}{}{\PackageError{access-migration}{Biography spacing patch failed}{}}
\def\@IEEEBIOskipN{1\baselineskip}
\expandafter\patchcmd\csname\string\IEEEbiography\endcsname
{\vskip \@IEEEBIOskipN plus 1fil minus 0\baselineskip}
{\vskip \@IEEEBIOskipN}{}{\PackageError{access-layout}{Biography glue patch failed}{}}
\makeatother
\newtheorem{proposition}{Proposition}
\newtheorem{theorem}[proposition]{Theorem}
\newtheorem{lemma}[proposition]{Lemma}
\newtheorem{corollary}[proposition]{Corollary}
\makeatletter\input{t1pcr.fd}\makeatother
\DeclareFontShape{T1}{pcr}{n}{n}{<->ssub*pcr/m/n}{}
\begin{document}
\history{}
\doi{}
\title{Exact Fusion of Softmax Attention Heads\\Can Require Exponentially Many Atoms}
\author{\uppercase{Hiroyasu Usami}\authorrefmark{1}, \uppercase{Keisuke Hara}\authorrefmark{1}, \uppercase{Ayato Tsuboi}\authorrefmark{1}, and \uppercase{Naohiko Matsuda}\authorrefmark{2}}
\address[1]{Department of Computer Science, Graduate School of Engineering, Chubu University, Kasugai, Aichi 487-8501, Japan}
\address[2]{Heat Transfer Research Department, Research \& Innovation Center, Mitsubishi Heavy Industries, Ltd., Takasago, Hyogo 676-8686, Japan}
\corresp{Corresponding author: Hiroyasu Usami (e-mail: usami@fsc.chubu.ac.jp).}
\tfootnote{}
\markboth{Usami et al.: Exact Softmax Head Fusion}{Usami et al.: Exact Softmax Head Fusion}
\begin{abstract}
Fusing independently normalized attention heads can increase the number of explicit key--value entries, or atoms, needed to preserve their output. We prove a sharp separation for $W_d(q)=\sum_{j=1}^d\sigma(q_j)$: its $d$ two-atom heads use $2d$ head--atom terms, whereas its minimum single-head representation has exactly $2^d$ atoms, even with arbitrary fixed replacement keys, scalar values, and positive masses. The same lower bound holds for every fixed uniform error below $1/2$ on $\mathbb R^d$, and the threshold is sharp. The proof uses exposed-key normal cones; exact equality on any nonempty open query domain inherits the bound. The resulting atom bound for a fixed head count is tight when that count divides $d$. A normalized family retains its exponential exact minimum while approaching a constant on fixed-radius query balls. For pretrained-model caches, a classical affine-rank bound and a certified instance exclude exact vector preservation at a specified atom count. Learned-cache comparisons on two disjoint banks of 64 contexts show budget-dependent reversals and calibration effects.
\end{abstract}
\titlepgskip=-18pt
\maketitle

\Needspace{6\baselineskip}
\section{Introduction}

Fusing separately normalized attention heads into a single softmax head changes the normalization structure. Within one readout, an atom is one explicit key--value entry with a fixed positive weight. In the constructed family, 12 two-atom heads use 24 head--atom terms, but any exactly equivalent single-head readout needs at least 4,096 atoms. Multiplying the head denominators produces a cache indexed by tuples of atoms. The question is whether jointly reoptimizing the replacement keys, values, and masses can reduce this product expansion.

We resolve this minimum-realization question for $W_d(q)=\sum_{j=1}^d\sigma(q_j)$. Its $d$ coordinate heads use two atoms each, while every single-head representation requires at least $2^d$ atoms, allowing arbitrary fixed real keys, scalar values, and positive constant masses. The obstruction is geometric: although $W_d$ has only $d+1$ distinct scalar limits along generic rays, a convex unique-winner region cannot combine directions across a coordinate jump. This forces $2^d$ exposed-key regions.

Keys, values, and masses are query-independent; queries range over $\mathbb R^d$.

\Needspace{18\baselineskip}
\noindent\textbf{Contributions.}
\begin{enumerate}
\item \textbf{Sharp atom cost.} We prove a matching exponential lower bound with a sharp global uniform-error threshold. A derived fixed-head budget is tight whenever the head count divides $d$.
\item \textbf{Exactness and approximation.} We distinguish exact preservation on open query domains from bounded-query approximation, and show that identical key geometry can have different exact atom requirements when values change.
\item \textbf{Finite-query capacity and learned approximation.} We apply a classical affine-rank bound to captured vector outputs and certify one native instance. We give separate and fused caches the same memory budget, fit both, and test how calibration-query placement changes their errors.
\end{enumerate}

\Needspace{6\baselineskip}
\section{Related work and positioning}

Jordan and Jacobs~\cite{Jordan1994} express a two-way softmax gate as a logistic ridge (their Eq. 2.7). We ask how small a flat, query-independent affine-logit readout can be for a specified function.

Haverbeck et al.~\cite{Haverbeck2026} analyze joint key--value compression by approximating a context measure under a query distribution, allowing new atoms and positive masses. Streaming memory bounds~\cite{Haris2025} and head-count or rank separations~\cite{Hsu2026,Yu2026,Amsel2025} address other targets and resources. Table~\ref{tab:prior} contrasts their interfaces with the minimum-realization question studied here.

Multi-query attention (MQA) shares one key--value cache across query heads, while grouped-query attention (GQA) shares a cache within each group~\cite{Shazeer2019,Ainslie2023}. Both preserve head-specific queries and softmax normalizations. The family $W_d$ itself can use two shared scalar KV positions, with keys $(0,1)$ and values $(0,1)$, coordinate queries $q_j$, and output summation. Thus KV sharing preserves the factored normalizers; fusion here replaces their separate distributions by one affine-logit softmax denominator.

Mixture-of-experts identifiability concerns equality of entire conditional distributions under irreducibility and kernel-independence conditions \cite{Jiang1999}. It does not directly determine the minimum number of constant scalar experts representing a prescribed mean. In particular, the stated binomial criterion requires $N\ge2m-1$ trials for $m$ experts, so its Bernoulli case $N=1$ does not cover $m\ge2$. The cube construction also repeats expert values for $d\ge2$. Our lower bound allows arbitrary fixed replacement keys, scalar values, and positive masses without an expert-density identifiability assumption.

\begin{table*}[t]
\caption{Closest results grouped by target, interface, and counted resource.}
\label{tab:prior}
\centering\footnotesize\setlength{\tabcolsep}{4pt}
\begin{tabular}{p{3.0cm}p{5.3cm}p{6.6cm}}
\toprule
Study & Target and interface & Guarantee or counted resource \\
\midrule
Jordan--Jacobs~\cite{Jordan1994} & Hierarchical conditional expert mixtures & Logistic gates and hierarchical products; construction background. \\
Haverbeck et al.~\cite{Haverbeck2026} & Joint KV replacement, including new atoms and positive masses & Query-distribution approximation risk; broader than support-subset eviction. \\
Haris--Onak~\cite{Haris2025} & Worst-case autoregressive streaming & Memory bits with probabilistic guarantees. \\
Hsu--Xu~\cite{Hsu2026} & Multi-task attention computation & Head count, dimension, precision, and threshold-degree tradeoffs. \\
Yu et al.~\cite{Yu2026} & Varying sequences, trainable encoders and post-attention networks & Transformer parameter approximation rates. \\
Amsel et al.~\cite{Amsel2025} & Retrieval with varying contexts and queries & Rank, head-count, and weight-norm barriers. \\
Jiang--Tanner~\cite{Jiang1999} & Full conditional expert densities under nondegeneracy & Parameter identifiability; not a scalar-mean minimum. \\
This work & Prescribed fixed scalar readout; arbitrary replacement keys, values, and positive masses & Minimum explicit atoms: exact on an open domain, or a sharp global uniform tolerance. \\
\bottomrule
\end{tabular}
\end{table*}

\Needspace{6\baselineskip}
\section{Readout model and fusion identity}

Let $\sigma(t)=(1+e^{-t})^{-1}$. For a positive integer $M$, consider
\[
B(q)=
\frac{\sum_{i=1}^{M}m_i e^{\langle q,k_i\rangle}v_i}
     {\sum_{i=1}^{M}m_i e^{\langle q,k_i\rangle}},
\qquad q\in\mathbb R^d,
\tag{1}
\]
where $k_i\in\mathbb R^d$, $v_i\in\mathbb R$, and $0<m_i<\infty$ are fixed. An atom is one indexed key-value-mass term in this expression. Identical keys can be merged by summing their masses and taking the mass-weighted value average. A constant positive mass is equivalent to a logit bias. Thus a lower bound for (1) also covers unit-mass replacement caches. A fixed scalar linear output projection can be absorbed into the values.

For head $h$, fix $P_h\in\mathbb R^{r_h\times d}$, keys $k_{hi}\in\mathbb R^{r_h}$, scalar values $v_{hi}$, and positive masses $m_{hi}$. With query $P_hq$, its output is
\[
A_h(q)=
\frac{\sum_{i=1}^{n_h}m_{hi}e^{\langle P_hq,k_{hi}\rangle}v_{hi}}
     {\sum_{i=1}^{n_h}m_{hi}e^{\langle P_hq,k_{hi}\rangle}}.
\]

\begin{proposition}[product expansion; background identity]
The sum $\sum_{h=1}^H A_h(q)$ has a representation of form (1) with at most $\prod_h n_h$ atoms. A tuple $I=(i_1,\ldots,i_H)$ has
\[
\begin{aligned}
k_I&=\sum_h P_h^{\mathsf T}k_{h i_h},\qquad m_I=\prod_h m_{h i_h},\\
v_I&=\sum_h v_{h i_h}.
\end{aligned}
\tag{2}
\]
\end{proposition}

\begin{proof}
Put the head outputs over the common denominator equal to the product of their denominators and expand. Each tuple contributes its product weight to the denominator. Its numerator coefficient is the same product weight multiplied by the sum of its selected values. All masses are positive. Coincident tuple keys may be merged, which only reduces the atom count.
\end{proof}

\Needspace{6\baselineskip}
\section{A sharp exponential atom bound}

Define
\[
W_d(q)=\sum_{j=1}^d \sigma(q_j),\qquad d\geq 1.
\tag{3}
\]
Each summand is a two-atom, unit-mass head with scalar keys $0,1$, values $0,1$, and coordinate query $q_j$.

\begin{theorem}[explicit-atom separation]
The minimum number of atoms in (1) representing $W_d$ on all of $\mathbb R^d$ is $2^d$. More strongly, if a representation (1) satisfies
\[
\sup_{q\in\mathbb R^d}|B(q)-W_d(q)|\leq\varepsilon
\qquad 0\leq\varepsilon<1/2,
\tag{4}
\]
then $M\geq 2^d$. In contrast, $d$ coordinate heads represent $W_d$ using $2d$ head-atom terms and a linear sum.
\end{theorem}

\begin{figure*}[t]
\centering
\includegraphics[width=\textwidth]{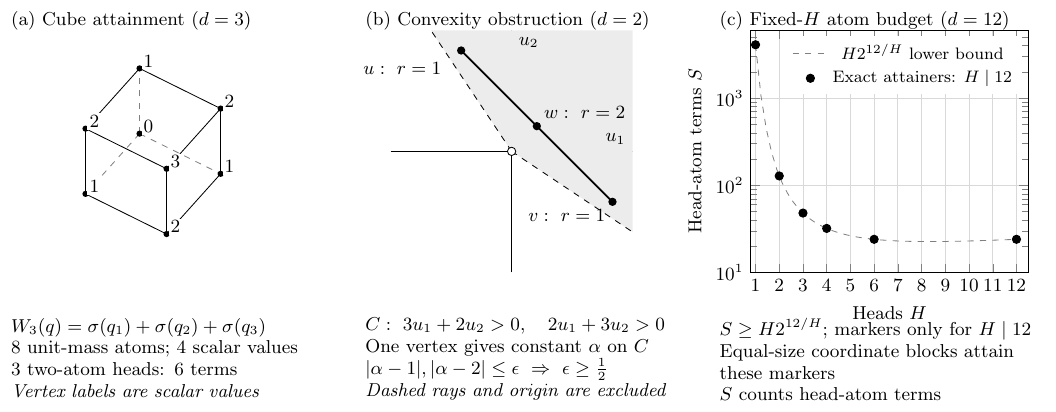}
\caption{Attainment, convexity obstruction, and fixed-head atom budgets. (a) Eight unit-mass cube atoms realize $W_3$; three two-atom heads use six head--atom terms. Vertex labels are scalar values, taking four distinct values. (b) The strict-winner cone for $k_0=0$ against keys $(-3,-2),(-2,-3)$ contains $u=(-1,2)$ and $v=(2,-1)$ with $r=1$, and their midpoint $w=(1/2,1/2)$ with $r=2$. One winning key gives one constant ray limit throughout its cone, requiring $\varepsilon\ge1/2$. Shading shows the clipped closure; dashed boundaries and the origin are excluded. (c) The curve is $H2^{12/H}$; filled markers attain it for $H\mid12$. Counts are head--atom terms.}
\label{fig:cones}
\end{figure*}

We use the following geometric observation; Figure~\ref{fig:cones}(b) illustrates its convexity obstruction. Panels (a) and (c) show the attaining cube construction and the fixed-head budget of Corollary~6, respectively.

\begin{lemma}[an open cone cannot hide a coordinate jump]
Let $C\subset\mathbb R^d$ be nonempty, open, and convex. Suppose that a constant $a$ satisfies
\[
|a-r(u)|\leq\varepsilon<1/2,\qquad
r(u)=\#\{j:u_j>0\},
\]
for every $u\in C$ with no zero coordinate. Then $C$ is contained in one open orthant.
\end{lemma}

\begin{proof}
If $C$ crosses a coordinate hyperplane $u_j=0$, convexity supplies an intersection and openness supplies a relatively open slice. Choose a slice point whose other coordinates are nonzero, and perturb it a sufficiently small distance in the two coordinate-normal directions. Both perturbed points remain in $C$, but their $r$-values differ by one. The assumed bounds imply $1\leq 2\varepsilon<1$, a contradiction. If $C$ contains a point on a coordinate hyperplane, openness gives the same contradiction. The argument also covers $d=1$.
\end{proof}

\begin{proof}[Proof of Theorem~2]
For the upper bound, take all cube keys $a\in\{0,1\}^d$, masses one, and values $v_a=\sum_j a_j$. The denominator factors as $\prod_j(1+e^{q_j})$. Summing the marginal expectation of each bit gives exactly $W_d$. These are also the tuples in Proposition 1.

For the lower bound, merge duplicate keys and let $K$ be the remaining finite key set. Each vertex $k$ of $\operatorname{conv}(K)$ has the nonempty, ambient-open, convex strict normal cone
\[
C_k=\bigcap_{k'\in K\setminus\{k\}}\{u:\langle u,k-k'\rangle>0\}.
\tag{5}
\]
This remains true if the key polytope has a lower-dimensional affine span. A singleton key set has $C_k=\mathbb R^d$.

For $u\in C_k$, the unique dominant exponential gives
\[
\lim_{t\to+\infty} B(tu)=v_k.
\]
For a direction with no zero coordinate,
\[
\lim_{t\to+\infty}W_d(tu)=r(u).
\tag{6}
\]
Condition (4) therefore implies $|v_k-r(u)|\leq\varepsilon$ on the generic directions of $C_k$. Lemma~3 places this cone inside a single orthant.

Every open orthant contains a direction avoiding all pairwise key-tie hyperplanes. In that direction a unique key maximizes the inner product, and it is an exposed vertex. A vertex cone cannot serve two orthants, so at least $2^d$ vertices, hence at least $2^d$ atoms, are necessary. This counts regions even though the ray limit takes only $d+1$ distinct values: equal values on disconnected orthants cannot be assigned to one open convex vertex cone. Exact equality is the case $\varepsilon=0$.
\end{proof}

\Needspace{11\baselineskip}
\section{Sharpness of the uniform-error threshold}

\begin{proposition}[boundary construction]
There is a $2^{d-1}$-atom representation with uniform error $1/2$ from $W_d$. For $d=1$, interpret it as a constant one-atom representation.
\end{proposition}

\begin{proof}
Drop the last coordinate and use
\[
B_{\mathrm{bdry}}(q)=
\sum_{j=1}^{d-1}\sigma(q_j)+\frac12.
\tag{7}
\]
The cube construction in $d-1$ coordinates uses $2^{d-1}$ unit-mass atoms; its values are the retained bit count plus $1/2$. Its error is $|1/2-\sigma(q_d)|$, whose supremum is $1/2$. Thus (4) cannot be extended to $\varepsilon=1/2$.

For every finite query the error in (7) is strictly less than $1/2$, while its supremum is $1/2$. Consequently pointwise strict error below $1/2$ does not imply the atom lower bound: a single fixed uniform tolerance strictly below $1/2$ is required.
\end{proof}

\begin{corollary}[exact preservation on an open domain]
Let $U\subset\mathbb R^d$ be nonempty and open. If a readout of form (1) equals $W_d$ throughout $U$, it has at least $2^d$ atoms. In particular, this applies to a bounded open box. It does not assert an approximation bound on a bounded domain.
\end{corollary}
\begin{proof}
For $a\in\{0,1\}^d$, write $|a|:=\sum_j a_j$. Put $D_W(q)=\prod_j(1+e^{q_j})$ and $N_W(q)=\sum_{a\in\{0,1\}^d}|a|e^{\langle q,a\rangle}$. Cross-multiplying the positive denominators gives the finite exponential sum
\[
E(q)=\sum_{i,a}m_i(v_i-|a|)e^{\langle q,k_i+a\rangle}.
\]
It vanishes on $U$. Fix $q_0\in U$ and any $q_*\in\mathbb R^d$. On the line $q_0+t(q_*-q_0)$, group equal exponents to write $E=\sum_{\ell=1}^L c_\ell e^{\lambda_\ell t}$ with distinct real $\lambda_\ell$. The line remains in $U$ for sufficiently small $|t|$, so derivatives of orders $0,\ldots,L-1$ at zero vanish. The resulting Vandermonde system implies every $c_\ell=0$. Hence $E(q_*)=0$, and positive denominators imply equality at every $q_*$. Apply Theorem~2.
\end{proof}

\begin{corollary}[fixed-head lower bound; sharp when $H\mid d$]
Suppose a sum of $H\ge1$ heads of the form preceding Proposition~1, with $n_h\ge1$ atoms in head $h$, equals $W_d$ on a nonempty open subset of $\mathbb R^d$, or approximates $W_d$ uniformly on all of $\mathbb R^d$ with fixed error $\varepsilon<1/2$. Then
\[
\prod_{h=1}^H n_h\ge2^d,\qquad
S:=\sum_{h=1}^Hn_h\ge H2^{d/H}.
\]
If $H$ divides $d$, the bound on $S$ is attained with total query width $d$.
\end{corollary}
\begin{proof}
Proposition~1 produces a single-head representation with at most $\prod_hn_h$ atoms. Theorem~2 or Corollary~5 requires at least $2^d$. The arithmetic--geometric mean inequality gives the bound on $S$. For attainment, partition the coordinates into $H$ blocks of size $d/H$ and apply the cube construction independently in each block. Every head has $2^{d/H}$ atoms, and summing the block outputs gives $W_d$ exactly.
\end{proof}
The bound allows arbitrary fixed linear query projections and constant heads, with a normalizer for each head and a linear sum of their outputs. It leaves the optimum over integer head sizes unspecified when $H$ does not divide $d$.

\Needspace{15\baselineskip}
\begin{proposition}[sharp one-atom comparison on a query ball]
Let $d\ge1$, $\tau>0$, $R>0$, and
\[
F_{d,\tau}(q)=\frac1d\sum_{j=1}^d\sigma(\tau q_j).
\]
On the origin-centered closed ball $\|q\|_2\le R$, the best one-atom approximation is the constant $1/2$, with error
\[
\begin{aligned}
E_1(d,\tau,R)
&:=\inf_{c\in\mathbb R}\sup_{\|q\|_2\le R}|F_{d,\tau}(q)-c|\\
&=\sigma(\tau R/\sqrt d)-\frac12.
\end{aligned}
\tag{8}
\]
Exact preservation on this ball still requires $2^d$ atoms.
\end{proposition}
The elementary proof is given in Appendix~A.

For $\tau=1/\sqrt d$, the cube representation has keys $a/\sqrt d$ of norm at most one and values $|a|/d\in[0,1]$. Its exact minimum on the ball is $2^d$, while
\[
E_1(d,1/\sqrt d,R)=\sigma(R/d)-\frac12\le\frac{R}{4d}.
\tag{9}
\]
At fixed $R$, this normalized family becomes nearly constant on the ball while retaining its exponential exact minimum (Table~\ref{tab:bounded}).

\begin{table}[t]
\caption{Closed-form theoretical comparison for $F_{d,1/\sqrt d}$ on $\|q\|_2\le1$. All cube keys have norm at most one and values lie in $[0,1]$. $M_{\rm exact}$ is the minimum single-head atom count; head terms count the coordinate construction. $E_1$ is the optimal one-atom uniform error. Values follow the closed forms of Proposition~7.}
\label{tab:bounded}
\centering\small\setlength{\tabcolsep}{4pt}
\begin{tabular}{rrrr}\toprule
$d$ & $M_{\rm exact}$ & Head terms & $E_1$ \\ \midrule
1 & 2 & 2 & 0.231059 \\
2 & 4 & 4 & 0.122459 \\
4 & 16 & 8 & 0.062177 \\
8 & 256 & 16 & 0.031209 \\
16 & 65536 & 32 & 0.015620 \\
\bottomrule\end{tabular}
\end{table}

\Needspace{6\baselineskip}
\section{Resource accounting and native-cache analysis}

The $2d$ quantity counts head--atom pairs, not sequence positions: several heads can read the same two positions. Rescaling the target to $W_d/d$ puts cube values in $[0,1]$ without changing the exact atom bound, but changes the uniform-error threshold to $1/(2d)$. Scaling cube keys by $1/\sqrt d$ changes the fixed-query target to $W_d(q/\sqrt d)$; a compensating query or temperature change must be counted.

Keep the cube key table and unit masses fixed. Values $v_a=a_1$ give $\sigma(q_1)$ after nuisance factors cancel, so the exact minimum is two atoms. Values $v_a=|a|/d$ give $W_d/d$, which requires $2^d$ atoms by Theorem~2. Both value sets lie in $[0,1]$, and the keys, rank, and key singular spectrum are identical. Thus key rank or key spectrum alone does not determine the minimum exact atom count.

For each context, normalized root mean squared error (NRMSE) is $\|\widehat Y-Y\|_F/\|Y\|_F$ with the uncentered target in the denominator; macro results give each context equal weight.

\subsection{Controlled fixed-cache readouts}

We use $d\in\{4,8,12\}$ and $R\in\{0.25,1,4\}$. For each $d$, a paired query bank contains 512 PCG64 uniform points in $[-1,1]^d$ (seed 20261101) and all $2^d$ sign corners, scaled by $R$. Four methods use the same generic, batched double-precision (FP64) NumPy softmax kernel: the cube head, two-coordinate block heads, coordinate heads, and a fixed two-atom approximation $d\sigma(d^{-1}\sum_jq_j)$. All K/V arrays are materialized; query reshaping and construction are outside kernel timing. Kernel timing details are retained in the supplement.

\begin{figure}[!t]
\centering
\includegraphics[width=0.98\columnwidth]{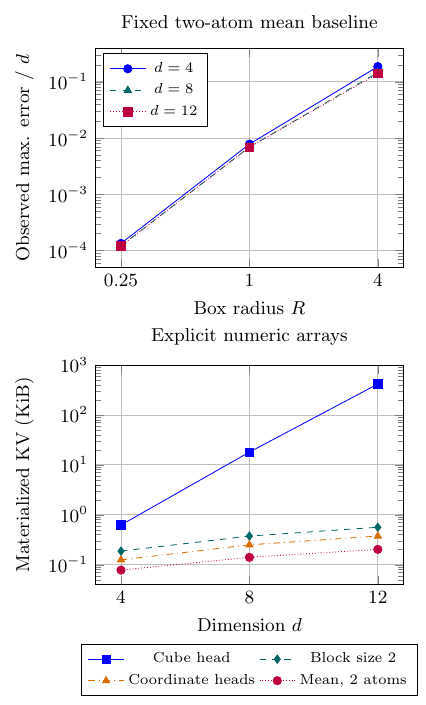}
\caption{Controlled readouts. Top: observed normalized maximum error for the fixed two-atom mean baseline, evaluated on paired finite query banks in boxes. Bottom: FP64 storage used by each method's K/V arrays. Exact-method errors across all banks are at most $7.04\times10^{-13}$.}
\label{fig:controlled}
\end{figure}

Across the 27 exact-method conditions, the maximum absolute error was $7.0344\times10^{-13}$. Figure~\ref{fig:controlled} shows the finite-bank approximation error and explicit array storage. Table~\ref{tab:controlled} reports the $d=12,R=1$ condition. At $d=12$, the cube ($H=1$), two-coordinate blocks ($H=6$), and coordinate heads ($H=12$) attain Corollary~6 with 4,096, 24, and 24 terms, respectively. Conversely, the approximate two-atom form is accurate on the smallest bank radius; the unbounded-query theorem supplies no bounded-query approximation barrier.

\begin{table}[t]
\caption{Controlled FP64 evaluation at $d=12$, $R=1$, with 4,608 queries. KV counts explicit array storage. The first three methods attain the fixed-head atom bound with $H=1,6,12$, respectively. The fixed mean is a prespecified two-atom approximation.}
\label{tab:controlled}
\centering\small
\setlength{\tabcolsep}{4pt}
\begin{tabular}{lrr}
\toprule
Method & Terms & KV (KiB) \\
\midrule
Cube head & 4096 & 416.000 \\
Block size 2 & 24 & 0.562 \\
Coordinate & 24 & 0.375 \\
Fixed mean & 2 & 0.203 \\
\bottomrule
\end{tabular}

\end{table}

\subsection{Pretrained-model targets and numerical identity}
We capture a pretrained model's head outputs and ask how many atoms are needed to preserve them. We then fit compressed caches and measure how approximation changes with the storage budget and calibration queries.

We use the public CODI GPT-2 checkpoint \nolinkurl{zen-E/CODI-gpt2}~\cite{CODI2025}. We run ordinary text inference on 64 label-free public BoolQ validation texts~\cite{BoolQAsset}, truncated to 96 GPT-2 tokens. We capture layer 0 and heads 0--3. Positions 0--15 supply a prefix cache, positions 16--47 calibrate compression, and positions 48 onward supply 2,932 held-out queries. The target renormalizes attention over that prefix. The vector analyses use concatenated head outputs before the model output projection.

Native 64-dimensional queries are divided by $8$ before concatenation; the fused dot product applies no further scaling. In the initial two-head scalar-projection diagnostic, full Cartesian fusion agrees with the separate reference to $8.88\times10^{-16}$ in FP64 and $4.77\times10^{-7}$ in FP32. At fused subset sizes 32 and 64, calibration top-mass pruning has larger macro error than uniform pruning. The supplement contains this diagnostic, its array accounting, and warm-kernel timings. The learned comparison uses trainable keys, values, and log masses.

For $H\in\{2,4\}$, the target is the concatenation of the first $H$ head-output vectors before the fixed output projection:
\[
Y_H(q)=\operatorname{concat}_{h=0}^{H-1}\bigl[\operatorname{softmax}(q_hK_h^\top)V_h\bigr]\in\mathbb R^{64H}.
\]

\subsection{Native-cache capacity at fixed KV payload}
An $n$-atom readout produces vectors in the affine hull of its $n$ values. Let $Y\in\mathbb R^{N\times D}$ stack target vectors from any fixed bank of $N$ queries, and let $C=I-\mathbf1\mathbf1^\top/N$. An $n$-atom readout has prediction matrix $B=PV$, where $P\mathbf1_n=\mathbf1_N$, so $\operatorname{rank}(CB)\le n-1$. Orthogonal centering and the low-rank approximation theorem~\cite{EckartYoung1936} give
\[
\frac{\|Y-B\|_F}{\|Y\|_F}\ge L_n(Y):=
\frac{\left(\sum_{j=n}^{\min(N,D)}\sigma_j(CY)^2\right)^{1/2}}{\|Y\|_F}.
\tag{10}
\]
We evaluate this floor numerically in FP64. It applies to the single-head class even with arbitrary replacement keys and values; softmax need not attain it.

A scalar target matrix has centered rank at most one, so the affine-rank floor vanishes for two or more atoms and cannot detect the exponential scalar minimum.

A second bank contains the next 64 texts in the same label-independent ordering, disjoint from the first bank. The same checkpoint, layer, two heads, and 16-position prefix define its targets.

Each original two-head cache occupies 16 KiB of FP32 key/value arrays. A 16-atom single head with 128-dimensional keys and values has the same KV payload. We additionally grant its 64 bytes of log masses free, enlarging the comparison class. On the 32 early calibration queries of each new context, (10) gives a macro NRMSE floor of 0.02486, ranging from 0.01773 to 0.03270 across all 64 contexts. The original separate heads realize their target exactly. Thus these native targets exhibit a finite-query fusion cost at equal dense KV payload. For the interval certificate, the captured FP32 query, key, and value entries define exact real constants, and the certified output is the corresponding real-exponential attention map. For one fixed native instance, an outward-rounded interval computation certifies a nonzero $17\times17$ minor of the augmented output matrix $[\mathbf1,Y]$. Its rank is therefore at least 17, excluding exact preservation by any 16-atom single head on those 17 queries. Appendix~B specifies the instance and verification.

\subsection{Learned approximation under an equal-byte budget}
On the first bank, we fit vector-valued records of form (1), with $v_i\in\mathbb R^{64H}$. The exported keys, values, and masses are fixed at evaluation.

The two-head condition is primary; the four-head condition is a prespecified nested comparison on the same capture. For $n\in\{4,8\}$, the separate heads receive $Hn$ block-local atoms and the single head receives $n$ full-dimensional atoms, at the same allocated memory budget (Table~\ref{tab:allocation}).

\begin{table*}[t]
\caption{Resource allocation in the equal-byte learned comparison. Both architectures retain the same number of key/value coefficients and receive the same FP32 storage budget. Normalizer count, atom count, coordinate access, and weight sharing vary together, allowing different training arithmetic at the same memory budget.}
\label{tab:allocation}\centering\small
\begin{tabular}{p{0.29\textwidth}p{0.31\textwidth}p{0.31\textwidth}}\toprule
Resource & Separate heads & Single head\\\midrule
Softmax normalizers & $H$ & $1$\\
Atoms; log masses & $Hn$; $Hn$ & $n$; $n$\\
Key and value width per atom & $64$ each & $64H$ each\\
Query and output coordinates & Fixed 64D block per head & All $64H$ coordinates per atom\\
Mixture-weight sharing & Within each 64D output block & Across all $64H$ output coordinates\\
Key coefficients; value coefficients & $64Hn$; $64Hn$ & $64Hn$; $64Hn$\\
FP32 array payload & $4(129Hn)$ bytes & $4(128Hn+n)$ bytes\\
Unused allocated padding & $0$ & $4n(H-1)$ bytes\\
Total allocated retention & $4(129Hn)$ bytes & $4(129Hn)$ bytes\\\bottomrule
\end{tabular}\end{table*}

Each cache uses 32 calibration queries, four fixed $k$-means++ initializations, and 512 Adam updates of keys and log masses, at learning rate $0.03$. Values are solved conditionally by ridge regression at each update and once after the final update. We scale the target using its calibration RMS and denote the result by $\widetilde Y$. For $D=64H$ and $N_c=32$, the objective shared by both architectures is
\[
J=\frac{\|AV-\widetilde Y\|_F^2}{N_cD}+\frac{10^{-6}\|V\|_F^2}{D}.
\]
Separate-head residuals and penalties are summed with the same denominator $D$. FP64 Cholesky solves give $(A^\top A+N_c10^{-6}I)V=A^\top\widetilde Y$; envelope gradients update the keys and log masses. Initial keys come from native-key clusters for separate heads and Cartesian-key clusters for the single head. The four-head initializer samples 512 fixed tuples; subsequent keys are unrestricted. The query scaling obtained only from calibration data is folded into the exported keys. We export the final caches in FP32. For each cache, the start with lowest exported calibration MSE is fixed before opening held-out assets.

\begin{table*}[t]
\caption{Learned compression with equal FP32 storage budgets. Each result uses all 64 prompts. Cal.: macro calibration NRMSE from FP32 exports; held: macro held-out NRMSE from FP32 records selected by calibration error. Lower is better. S/F denote separate/single-head fits. Separate/single allocations retain $Hn$/$n$ total atoms. Fits use 512 updates and four starts; Table~\ref{tab:allocation} details the resource allocation.}
\label{tab:learned}\centering\small
\begin{tabular}{rrrrrrrr}\toprule
$H$ & $n$ per head & KiB each & Cal. S & Cal. F & Held S & Held F & S lower\\\midrule
2 & 4 & 4.031 & 0.1901 & 0.2355 & 0.4948 & 0.4467 & 32/64 \\
2 & 8 & 8.062 & 0.0573 & 0.1179 & 0.3363 & 0.5269 & 62/64 \\
4 & 4 & 8.062 & 0.1550 & 0.2845 & 0.4855 & 0.5135 & 42/64 \\
4 & 8 & 16.125 & 0.0463 & 0.2071 & 0.3657 & 0.4896 & 60/64 \\
\bottomrule\end{tabular}
\end{table*}

\paragraph{Results.}
The primary two-head comparison does not meet the prespecified two-budget superiority criterion: at $H=2,n=4$, the single-head macro error is lower (0.4467 versus 0.4948); at $n=8$, separate heads are lower (0.3363 versus 0.5269). Separate heads have lower errors on 32/64 and 62/64 contexts, respectively. The median single-minus-separate differences are $-0.0011$ and $0.1179$. At $H=4$, separate heads are lower at both budgets, on 42/64 and 60/64 contexts.

Calibration error decreases with capacity in both classes, whereas the fitted single head's held-out error increases in the primary condition (Table~\ref{tab:learned}). Across starts, the largest NRMSE range within a context is 1.8158 for separate heads at $H=2,n=4$, and 1.1793 for the single head at $H=2,n=8$. Macro rank floors are 0.2539 and 0.1265 for two heads, and 0.2559 and 0.1495 for four heads. Every selected separate fit lies above its corresponding single-head floor, so this native bank supplies no separation certified by that bound.

\subsection{Calibration-query coverage}
We compare calibration-query placement on the second bank. Both policies retain 32 calibration queries and share the final 16 queries as their held-out set. The early policy uses positions 16--47. For a text of $T$ tokens, the distributed policy uses positions $16+\lfloor j(T-33)/31\rfloor$, $j=0,\ldots,31$, before the held-out interval $[T-16,T)$. Here $67\le T\le96$. Keys, values, fitting steps, starts, and retention budgets follow the same procedure; all eight selectors are fixed before evaluating either policy. Changing the calibration queries also changes the scaling and initialization used by the fitting procedure.

\begin{table}[t]\centering\small
\caption{Calibration-policy follow-up on 64 unused contexts. Held-out NRMSE is averaged equally across prompts on the common bank of final-16 queries. S/F: separate/single-head fits.}
\label{tab:coverage}
\begin{tabular}{rrrrrr}\toprule
& & \multicolumn{2}{c}{Early} & \multicolumn{2}{c}{Distributed}\\
$n$ & KiB & S & F & S & F\\\midrule
4 & 4.031 & 0.4967 & 0.4809 & 0.4717 & 0.4711 \\
8 & 8.062 & 0.3644 & 0.5734 & 0.3515 & 0.4817 \\
\bottomrule\end{tabular}\end{table}

\begin{figure*}[t]
\centering
\includegraphics[width=0.98\textwidth]{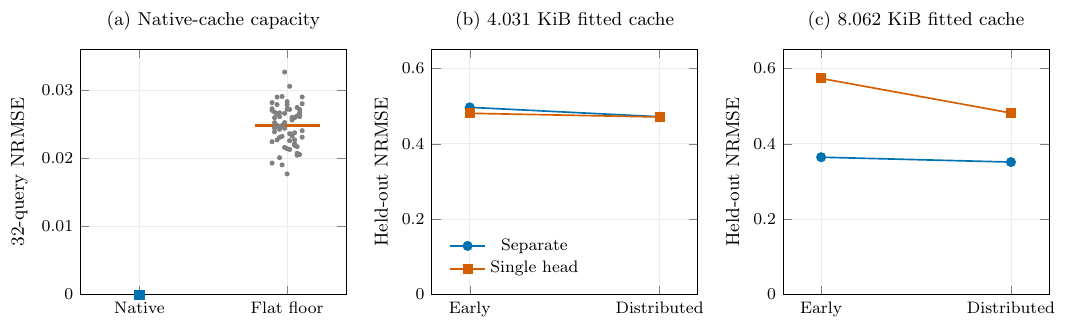}
\caption{Native capacity and calibration on the second, disjoint 64-context bank. (a) The original two-head cache realizes its target with a 16 KiB KV payload. Gray points are classical affine-rank error floors for arbitrary 16-atom single heads with the same KV payload and free masses, evaluated on 32 early queries per context; the red bar is their mean. (b,c) Macro error on the common final-16-query bank when the 32 calibration queries are early or distributed. Both architectures improve at both fitted-cache budgets.}
\label{fig:learned}
\end{figure*}

Distributed calibration lowers macro NRMSE in all four planned contrasts: by 0.0250 and 0.0098 for separate and single heads at 4.031 KiB, and by 0.0129 and 0.0918 at 8.062 KiB. At the smaller budget, the two errors are close (0.4717 and 0.4711); at the larger budget, separate heads are lower (0.3515 versus 0.4817). The worst per-context separate-head error at the smaller budget decreases from 1.5183 to 0.8585. The mean nearest-calibration cosine distance of held-out queries decreases from 0.4916 to 0.4240 in the fixed native query coordinates.

\Needspace{6\baselineskip}
\section{Implications for KV compression}

A factored evaluator computes $W_d$ with $O(d)$ sigmoid evaluations; a single normalized exponential sum of form (1) needs $2^d$ atoms.

The scalar obstruction also transfers to vector targets through a fixed linear functional. If $F:\mathbb R^d\to\mathbb R^p$ satisfies $\ell(F(q))=W_d(q)$ for a fixed nonzero $\ell$, applying $\ell$ to vector values preserves the keys, positive masses, and atom count. Theorem~2 and Corollary~5 therefore give the same exact lower bound. A global uniform vector error $\varepsilon$ also gives $M\ge2^d$ whenever $\|\ell\|_*\varepsilon<1/2$.

Key/value dimensions, precision, normalizers, and evaluation kernels determine how atoms translate into storage and time. Table~\ref{tab:allocation} records the matched storage resources and the architectural differences in the learned comparison.

\Needspace{6\baselineskip}
\section{Conclusion}

Fusing softmax heads can require exponentially many atoms even after all replacement keys, scalar values, and masses are reoptimized. The minimum single-head representation of $W_d$ has $2^d$ atoms, with a sharp all-query uniform-error threshold of $1/2$. For $H$ separately normalized heads, the total atom count is at least $H2^{d/H}$, attained when $H$ divides $d$. The normalized family becomes nearly constant on a bounded query ball while retaining its exact minimum.

For one native vector target, the certificate rules out exact single-head preservation at the original dense KV payload. Learned-cache rankings vary with storage budget, while distributed calibration reduces macro error in both allocations.

\Needspace{5\baselineskip}
\section*{Code and data availability}

\begingroup\raggedright
The reproducibility materials are maintained at \url{https://github.com/usamihiroyasu/attention-head-fusion}. The same files are supplied as supplementary material. They include diagnostic and learned-cache implementations, frozen protocols, both query banks with upstream attribution, complete results and all-start diagnostics, and environment and asset identifiers. Model weights are retrieved from the cited pinned repository rather than redistributed.\par\smallskip
The checkpoint revision is \nolinkurl{fd641b3d3edc59e4f534b55588e906588c9e36bb}. All 251 checkpoint entries have the expected keys and shapes. Every loaded tensor equals its checkpoint value after conversion to FP32. The dataset Parquet file has SHA-256 prefix \nolinkurl{52355d11524b4b87}; the supplement records the full digest and all input hashes. All 36 method--bank conditions completed. All eight conditions completed for all 64 contexts.\par
\endgroup

\appendices
\Needspace{6\baselineskip}
\section{Proof of Proposition 7}
\begin{proof}
Every one-atom readout is constant. Put $f(x)=\sigma(\tau x)-1/2$ for $x\ge0$. This function is increasing and concave. Odd symmetry of $\sigma(t)-1/2$, Jensen's inequality, and Cauchy--Schwarz give
\[
\begin{aligned}
|F_{d,\tau}(q)-1/2|
&\le\frac1d\sum_jf(|q_j|)\\
&\le f\!\left(\frac{\|q\|_1}{d}\right)
\le f(R/\sqrt d).
\end{aligned}
\]
The opposite points $q=\pm(R/\sqrt d)\boldsymbol{1}$ attain the two symmetric extrema. Their output separation is twice the displayed error, so no constant does better. Exact preservation follows from Corollary~5 after the invertible query scaling by $\tau$ and output scaling by $d$; the ball has nonempty interior.
\end{proof}

\section{A certified native-cache instance}
Take context 0 of the second bank (BoolQ validation row 842), its first 17 early queries (positions 16--32), and the fixed prefix from positions 0--15. Treat the captured FP32 query, key, and value entries as exact real constants. Form a $17\times17$ matrix $A$ with a constant-one column, output coordinates 0--14 of head 0, and coordinate 0 of head 1.

The certificate encloses every dot product, exponential, positive normalization, and output using 80-digit decimal arithmetic. Correctly rounded endpoint exponentials~\cite{PythonDecimal} are expanded outwards using \texttt{next\_minus} and \texttt{next\_plus} in the same 80-digit context; the remaining interval operations use directed rounding. Interval elimination gives
\[
\begin{split}
3.3110337468158202\times10^{-30}<\det A\\
<3.3110337468158203\times10^{-30}.
\end{split}
\]
An independent check uses a proposed numerical inverse $R$ only as an exact rational preconditioner. Exact fraction arithmetic shows that every matrix in the stored entry enclosure has $RA$ strictly row diagonally dominant, with all dominance margins exceeding $0.9999999993$. Hence it is nonsingular. The input arrays, entry enclosures, preconditioner, and two verification programs accompany the supplement.

For an $M$-atom flat readout, $B=PV$ and $P\mathbf1_M=\mathbf1$, so $[\mathbf1,B]=P[\mathbf1_M,V]$ has rank at most $M$. The certified rank 17 excludes $M\le16$ for this instance.

\noindent\begin{minipage}[t]{\columnwidth}
\begin{IEEEbiography}[{\includegraphics[width=1in,height=1.25in,keepaspectratio,trim=325bp 94bp 33bp 40bp,clip]{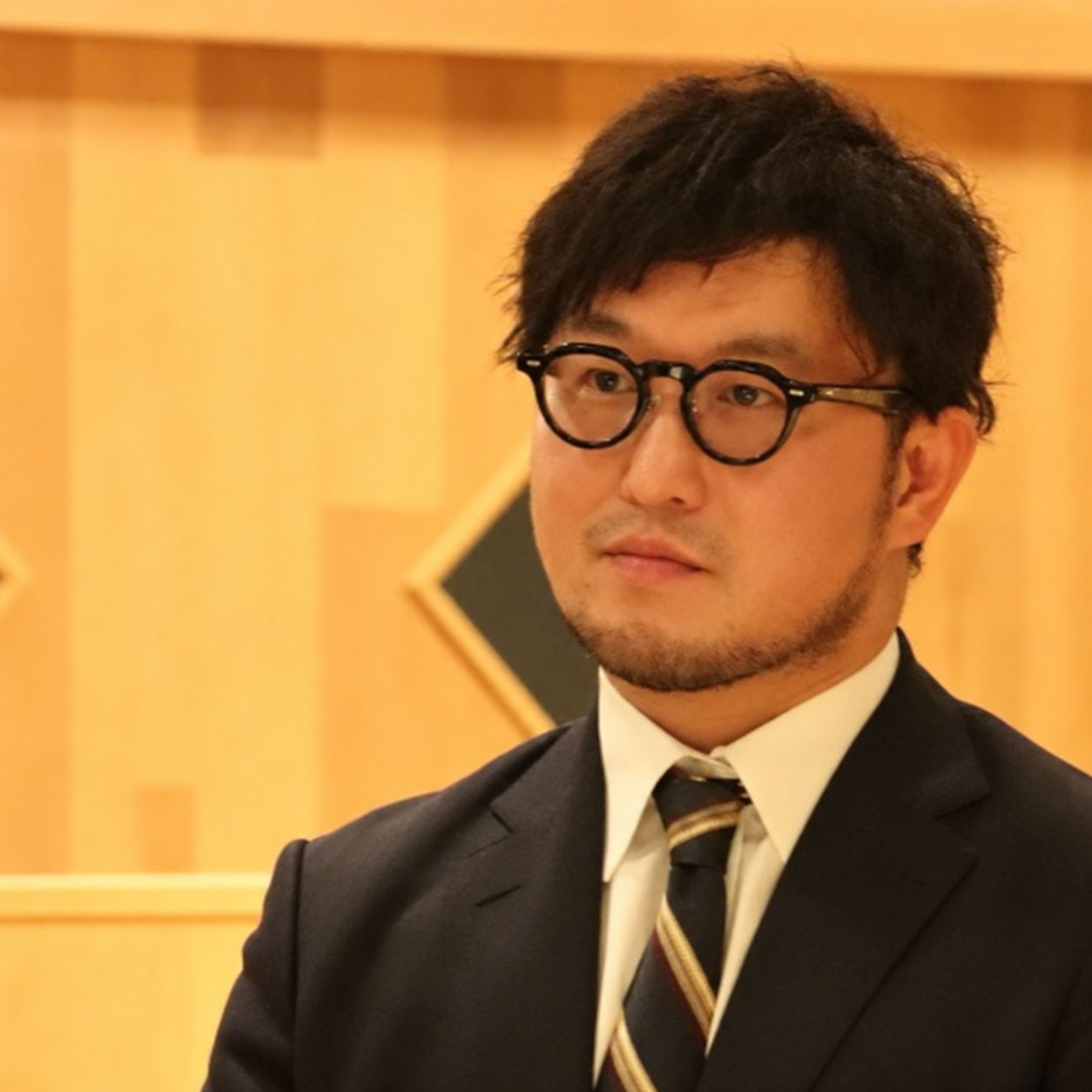}}]{Hiroyasu Usami}
received the Ph.D. degree in engineering from Chubu University in 2019. He is a Senior Assistant Professor at Chubu University and a Researcher at Aichi Medical University. His research focuses on computer vision, machine learning, and medical image analysis. He serves on the Editorial Board of \emph{Scientific Reports} and the editorial committee of the \emph{IPSJ Journal/Journal of Information Processing}.
\end{IEEEbiography}
\end{minipage}\par\vspace{5pt}

\noindent\begin{minipage}[t]{\columnwidth}
\begin{IEEEbiography}[{\includegraphics[width=1in,height=1.25in,keepaspectratio]{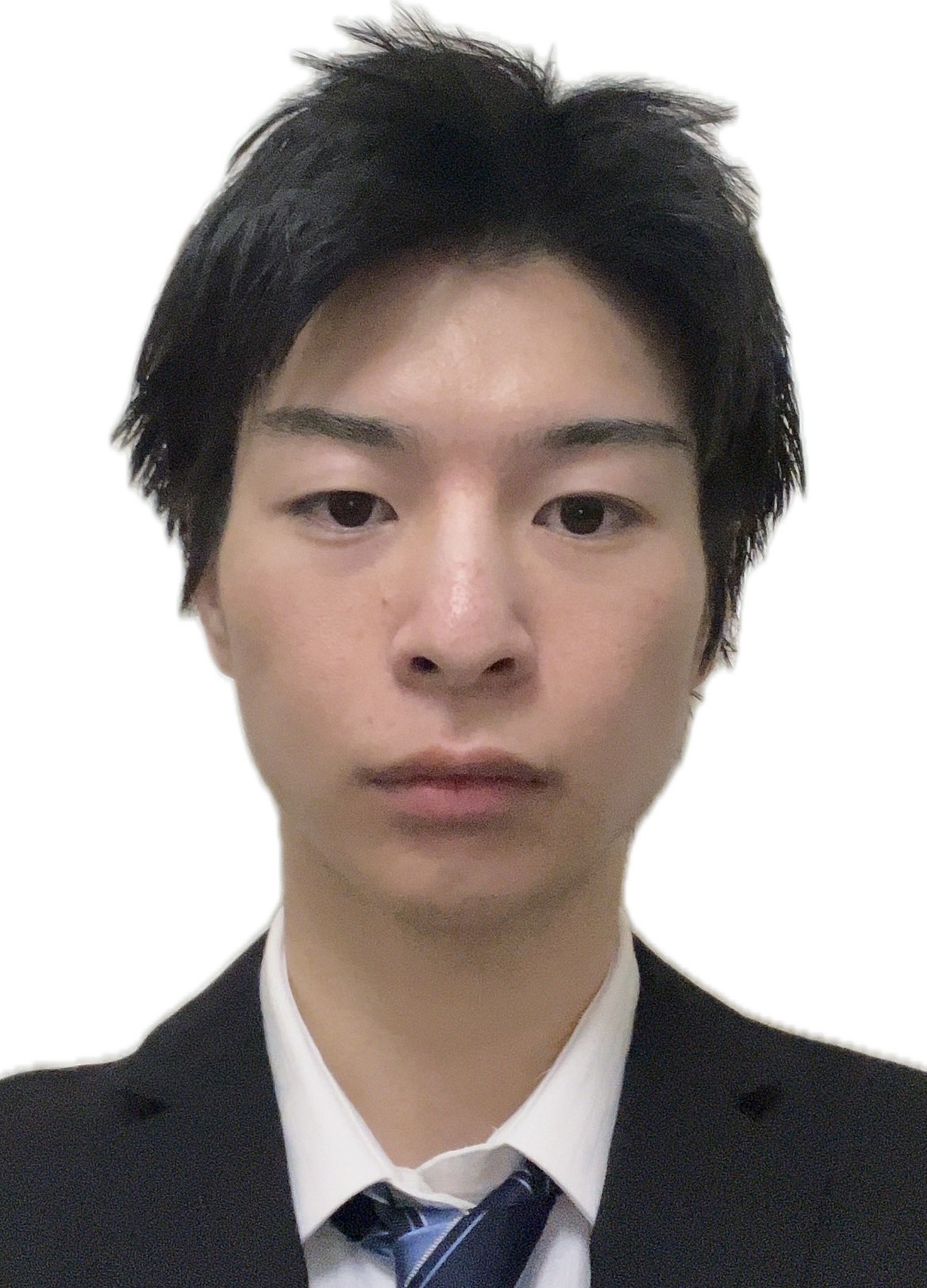}}]{Keisuke Hara}
received the B.E. degree in computer science from Chubu University, Kasugai, Japan, in 2026. He is currently pursuing the M.E. degree in computer science at the Graduate School of Engineering, Chubu University. His research interests include computer vision, image decomposition, shape from shading, 3-D reconstruction, depth estimation, and medical image computing.
\end{IEEEbiography}
\end{minipage}\par\vspace{5pt}
\newpage

\noindent\begin{minipage}[t]{\columnwidth}
\begin{IEEEbiography}[{\includegraphics[width=1in,height=1.25in,keepaspectratio]{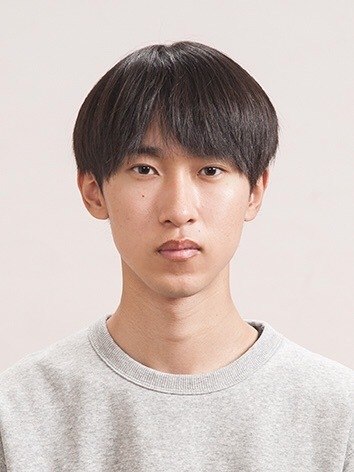}}]{Ayato Tsuboi}
received the B.E. degree in computer science from Chubu University, Kasugai, Japan, in 2026. He is currently pursuing the M.E. degree in computer science at the Graduate School of Engineering, Chubu University. His research interests include computer vision, image decomposition, inverse rendering, 3-D reconstruction, shape from polarization, and medical image processing.
\end{IEEEbiography}
\end{minipage}\par\vspace{5pt}

\noindent\begin{minipage}[t]{\columnwidth}
\begin{IEEEbiography}[{\includegraphics[width=1in,height=1.25in,keepaspectratio]{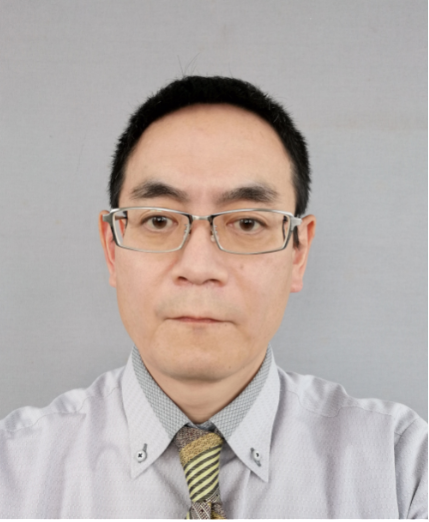}}]{Naohiko Matsuda}
is with the Heat Transfer Research Department, Research \& Innovation Center, Mitsubishi Heavy Industries, Ltd., Japan. His current work includes the application of large language models to engineering analysis and energy-system modelling.
\end{IEEEbiography}
\EOD
\end{minipage}
\end{document}